\pdfoutput=1

\documentclass[11pt]{article}

\usepackage[final]{coling}

\usepackage{times}
\usepackage{latexsym}

\usepackage[T1]{fontenc}

\usepackage[utf8]{inputenc}

\usepackage{microtype}

\usepackage{inconsolata}

\usepackage{graphicx}
\usepackage{tabularx}
\usepackage{enumitem}
\usepackage{xspace}
\usepackage{booktabs}

\newif\ifcomments
\ifcomments
\newcommand{\draftcomment}[3]{{\color{#2}[\textsc{#1} #3]}}
\else
\newcommand{\draftcomment}[3]{}
\fi

\newcommand{\rob}[1]{\draftcomment{Rob:}{purple}{#1}}
\newcommand{\roberto}[1]{\draftcomment{Roberto:}{orange}{#1}}

\newcommand{\aoife}[1]{\draftcomment{Aoife:}{violet}{#1}}
\newcommand{\neema}[1]{\draftcomment{Neema:}{cyan}{#1}}

\definecolor{KenFlagRed}{HTML}{EA9B00}
\definecolor{KenFlagGreen}{HTML}{006600}

\newcommand{\datasetname}{Uchaguzi-2022\xspace}
\newcommand{\spaceauthor}{\hspace{0.3cm}}

\title{\datasetname: A Dataset of Citizen Reports on the 2022 Kenyan Election}

\author{
 \textbf{Roberto Mondini\textsuperscript{\textdagger}}
 \spaceauthor
 \textbf{Neema Kotonya\textsuperscript{\textdagger}}
 \spaceauthor
 \textbf{Robert Logan\textsuperscript{\textdagger}}
  \spaceauthor
 \textbf{Elizabeth Olson\textsuperscript{\textdagger}}
\\
 \textbf{Angela Oduor Lungati\textsuperscript{\textdaggerdbl}}
  \spaceauthor
 \textbf{Daniel Duke Odongo\textsuperscript{\textdaggerdbl}}
  \spaceauthor
 \textbf{Tim Ombasa\textsuperscript{\textdaggerdbl}}
\\
 \textbf{Hemank Lamba\textsuperscript{\textdagger}}
  \spaceauthor
 \textbf{Aoife Cahill\textsuperscript{\textdagger}}
  \spaceauthor
 \textbf{Joel Tetreault\textsuperscript{\textdagger}}
  \spaceauthor
 \textbf{Alejandro Jaimes\textsuperscript{\textdagger}}
\\
 \textsuperscript{\textdagger}Dataminr Inc.
 \textsuperscript{\textdaggerdbl}Ushahidi
\\
   \texttt{\{rmondini,neema.kotonya,rlogan,elizabeth.olson,}\\
   \texttt{hlamba,acahill,jtetreault,ajaimes\}@dataminr.com}
 \\
   \texttt{\{angela,daniel,tim.ombasa\}@ushahidi.com}
}

\begin{document}
\maketitle
\begin{abstract}
Online reporting platforms have enabled citizens around the world to collectively share their opinions and report in real time on events impacting their local communities.
Systematically organizing (e.g., categorizing by attributes) and geotagging large amounts of crowdsourced information is crucial to ensuring that accurate and meaningful insights can be drawn from this data and used by policy makers to bring about positive change.
These tasks, however, typically require extensive manual annotation efforts.
In this paper we present \datasetname, a dataset of 14k categorized and geotagged citizen reports related to the 2022 Kenyan General Election containing mentions of election-related issues such as official misconduct, vote count irregularities, and acts of violence.
We use this dataset to investigate whether language models can assist in scalably categorizing and geotagging reports, thus highlighting its potential application in the AI for Social Good space.\footnote{Dataset and code available at: \url{https://github.ushahidi.org/uchaguzi-ai/}}
\end{abstract}

\section{Introduction}

Citizen journalism (i.e., non-professional reporting disseminated on social media and dedicated websites) plays an increasingly important role in enabling the circulation of opinions on topics such as elections, as well as in exposing electoral violence and interference~\cite{moyo_citizen_2009,ajao_citizen_2017}.
To ensure credibility, it is incumbent on social media platforms and websites to verify citizen-submitted reports, since unmoderated content can exacerbate the spread of misinformation~\cite{ndlela2020social}.
As reports are often issued via social media posts, it is also crucial for platforms to be able to \emph{categorize} (e.g., by topic) and \emph{geotag} (i.e., add geographic metadata to) posts to help readers understand how the reported events are impacting their communities.
Furthermore, the systematic organization of large amounts of crowdsourced information enables reporting agencies to draw meaningful insights from this data and share these with the relevant policy makers to enact positive changes~\cite{shayo2021citizen}.
Unfortunately, categorizing and geotagging large amounts of data typically requires substantial manual effort, often led by volunteers~\cite{aarvik2015uchaguzi,shayo2021citizen}, severely delaying and limiting the publication and circulation of information.

\begin{figure}[!t]
    \centering
    \includegraphics{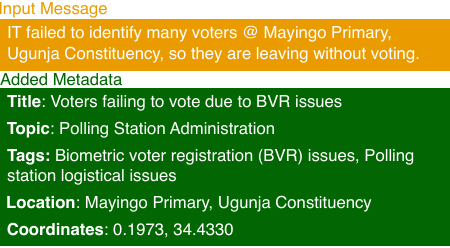}
    \caption{
        \textbf{Annotated report example}.
        Each input message \textit{(top)} is categorized and geolocated \textit{(bottom)}.
    }
    \label{fig:data-example}
\end{figure}
In this paper we introduce \datasetname
, a dataset of 14k reports pertaining to the 2022 Kenyan General Election
that have been categorized and geotagged (Figure \ref{fig:data-example}).
Unlike existing social media-based election datasets, e.g., \citet{wang_system_2012}, \citet{schmidt_sentiment_2022}, that focus on studying electorate sentiment and community structure, \datasetname contains reports of election interference issues such as official misconduct, vote count irregularities, and acts of violence.
Through this dataset we explore NLP methods to automate report categorization and geotagging.
We provide benchmark results for both encoder-only language models~\cite{conneau-etal-2020-unsupervised} trained using fine-tuning~\cite{devlin-etal-2019-bert}, and decoder-only language models~\cite{jiang2024mixtral} trained using few-shot in-context learning~\cite{brown2020language}.
By alleviating the manual efforts to complete these tasks, systems trained on this dataset will allow reporting organizations to focus on extracting meaningful insights from this structured data and help inform decisions that can positively impact the reporting citizens and their communities.

In summary, our contributions are as follows:
\begin{enumerate}[nosep,leftmargin=*]
    \item We present \datasetname, a dataset of categorized and geotagged citizen reports on the 2022 Kenyan General Election. To our knowledge, this is the first dataset of citizen-contributed election issues in the African continent that contains this associated metadata.
    \item We use this dataset to benchmark models for assisting with categorization and geotagging, demonstrating the potential application of this data for broader AI for Social Good efforts. Our results show that few-shot models are competitive with fully fine-tuned models.
\end{enumerate}

\section{The \datasetname Dataset}
\label{sec:dataset}

\subsection{Overview}
\label{sec:dataset-overview}
\datasetname
contains 14,169 citizen reports related to the Kenyan General Elections held on August 9, 2022 and submitted to the Uchaguzi platform\footnote{https://uchaguzi.or.ke/} between June 27 and August 29, 2022.
This platform, developed and maintained by Ushahidi\footnote{https://www.ushahidi.com/}, enables citizens to share their views and report on election-related events through SMS messages, X (formerly Twitter) posts, and questionnaires initiated via USSD and WhatsApp.
The submitted reports are manually reviewed and annotated by a team of Ushahidi volunteers, who additionally can create their own reports from public social media posts.
Volunteers are fluent in both English and Swahili and participate in training sessions prior to undertaking the annotation tasks \cite{ushahidi-uchaguzi-teams-2022}.
After the volunteers' review, reports are surfaced to the public and reporting agencies through the Uchaguzi website \cite{ushahidi-uchaguzi-2022}.

\begin{table}[!t]
    \centering
    \small
    \begin{tabular}{lrr}
       \toprule
       \textbf{Topic}  &  \textbf{Count} \\
       \midrule
       Opinions & 9,441 \\ 
       Media Reports & 1,736 \\ 
       Positive Events & 1,256 \\ 
       Counting and Results & 755 \\ 
       Security Issues & 302 \\ 
       Voting Issues & 285 \\ 
       Political Rallies & 134 \\ 
       Polling Station Administration & 123 \\ 
       Staffing Issues & 76 \\ 
       Irrelevant Report & 61 \\ 
       \bottomrule
    \end{tabular}
    \caption{
        \textbf{List of topics and their prevalence}.
    }
    \label{tab:topic-dist}
\end{table}


\begin{table}[!t]
    \small
    \centering
    \begin{tabular}{lr}
        \toprule
        \textbf{Tag}  &  \textbf{Count} \\
        \midrule
        Biometric voter registration (BVR) issues & 38 \\
        Polling station logistical issues & 38 \\
        Low voter turn out & 23 \\
        Missing/inadequate voting materials & 14 \\
        Ballot box irregularities & 9 \\
        Polling station closed before voting concluded & 5 \\
        Campaign material in polling station & 3 \\
        High voter turn out & 2 \\
        \bottomrule
    \end{tabular}
    \caption{
        \textbf{Tags for the \emph{Polling Station Administration} topic and their prevalence}.
    }
    \label{tab:tag-example}
\end{table}

Volunteers \emph{categorize} incoming reports in two ways: 1) by assigning a \emph{topic}, and 2) by assigning topic-specific \emph{tags}.
There are ten topics (Table~\ref{tab:topic-dist}), while tags provide further fine-grained categorization.
The set of applicable tags is dependent on the topic selected, is optional, and multiple tags can be assigned to a report.
For illustration, the tag set for the {\it Polling Station Administration} topic is listed in Table~\ref{tab:tag-example}, and the tag sets for the remaining topics as well as examples for each topic are provided in Appendix~\ref{sec:appendix-tag-assignment-categories}.

As tags are added, volunteers also give reports a \emph{title}, i.e., a short summary of the content of the report, and \emph{geotags}.
Geotags are provided in the form of \emph{coordinates} (latitude and longitude) as well as an optional \emph{location} name, and are determined by looking for location mentions in the source message.
If no location is mentioned, the report is associated to a default location in the center of Nairobi so that it may still be appear on the Uchaguzi website's map.

Finally, a separate team of volunteers is responsible for reviewing the annotated reports, ensuring the correct metadata were added, and for publishing them to the Uchaguzi website \cite{ushahidi-uchaguzi-publishing-2022}.
Figure~\ref{fig:data-example} shows an example of a fully annotated report. 

\subsection{Preprocessing}

The 14,169 reports in the \datasetname dataset represent the subset of all incoming reports reviewed and annotated by volunteers.
An additional 86k reports were received by the platform, but were left unannotated due to a lack of resources, demonstrating the need for an automated approach to address the scale of the problem.
To create the annotated dataset, we exclude reports without an assigned topic (86k reports).
We focus on text-based reports submitted via SMS, X (formerly Twitter), and public social media posts (as they share similar formats), and further exclude around 1,600 questionnaires (since metadata is added deterministically for this type of report).
The annotated dataset along with the code are available at: \url{https://github.ushahidi.org/uchaguzi-ai/}.

\begin{figure}[t]
    \centering
    \includegraphics{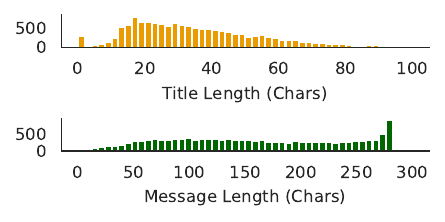}
    \caption{{\bf Distributions of title and message lengths.}}
    \label{fig:report-text-title-len-dist}
\end{figure}

\begin{table}[!t]
    \centering
    \small
    \begin{tabular}{lr}
        \toprule
       \textbf{Field}  &  \textbf{\% Non-empty}\\
       \midrule
       \texttt{title} & $97.9$\\
       \texttt{topic} & $100.0$\\
       \texttt{tags} & $85.1$\\
       \texttt{location} & $2.4$\\
       \texttt{coordinates} & $98.4$\\
        \bottomrule
    \end{tabular}
    \caption{
        {\bf Non-empty fields in \datasetname.}
    }
    \label{tab:dataset_non_empty_fields}
\end{table}

\subsection{Analysis}

In this section, we include analyses of annotation coverage, lexical content, geographic distribution, and temporal trends within \datasetname.

\paragraph{Annotation Coverage}
Each report in \datasetname has an assigned topic, but not all reports were further annotated with title, tags, or geotagging information (Table~\ref{tab:dataset_non_empty_fields}).
In particular, while \textit{coordinates} are provided for the majority of reports (98\%), there is relatively low coverage of the \textit{location} text field (2.4\%). Although the reports may not mention any location, in \S\ref{sec:geotagging-results} we find that this low coverage is also in part due to annotation incompleteness.
The distribution of topic labels is shown in Table~\ref{tab:topic-dist}.
The most frequent topic, \textit{Opinions}, was assigned to 9,441 reports ($66.6\%$), whereas the least frequent topic, \textit{Irrelevant Report}, was assigned to 61 reports ($0.4\%$) and generally indicated irrelevant SMS messages received by the platform.

\paragraph{Lexical Content}
In Figure \ref{fig:report-text-title-len-dist} we present the report title and message length distributions.
The spike in message lengths around 280 characters likely corresponds to the (then) Twitter length limit.
Additionally, as many languages are spoken in Kenya~\cite{muaka-language-2011}, we study the language distribution in the reports with \texttt{fasttext}~\cite{joulin2016bag}.
According to this model, $98.0\%$ of reports are in English, $1.3\%$ are in Swahili, and the remaining $0.7\%$ are other languages.
Among the reports detected as English, we observe the presence of English-Swahili code-switching and show examples in Table \ref{tab:lang-examples}.
\aoife{anything else to say here about how much of the 98\% is likely to contain at least some non-English?}
\roberto{No, we did not do any further analyses on code-switching as we agreed those were outside the scope of this paper and best left for future work.}

\begin{figure}[t]
    \centering
    \includegraphics{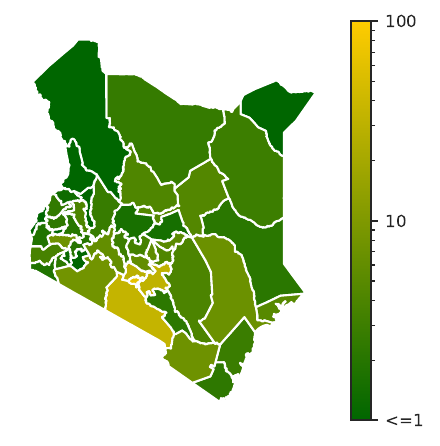}
    \caption{{\bf Reports per capita.} Scale is per 100K citizens.}
    \label{fig:geo-dist}
\end{figure}

\begin{table*}[t]
    \centering
    \small
    \begin{tabular}{p{7.0cm}p{7.0cm}}
       \toprule
       \textbf{Text} &  \textbf{Translation} \\
       \midrule
       So why are we going to the ballot if it’s so obvious to you that your candidate has already won? Anyway sisi tunajua kura ni 9/8/22, wa Kenya ndio wataamua na IEBC ndio watatangaza mshindi sio ng’ombe za AZIMIO. & So why are we going to the ballot if it's so obvious to you that your candidate has already won? Anyway, we know the vote is 9/8/22, Kenyans will decide and IEBC will announce the winner, not the cows of AZIMIO.\\
       \midrule
       Cherera alikua chief of staff wa Joho. waliwekwa kwa commission ile time ya handshake when there was a debate about IEBC commissioners Quorum. She is out to pay her dues \#kenyaelections2022 Serena kisumu Ledama & Cherera became Joho's chief of staff. they were appointed to the commission at the time of the handshake when there was a debate about IEBC commissioners quorum. She is out to pay her dues \#kenyaelections2022 Serena Kisumu Ledama\\
       \midrule
       DCI na IEBC nao ni kama wanatupiga kipindi, tutaona mambo before election day. & DCI and IEBC are like they are giving us a show, we will see things before election day.\\
       \bottomrule
    \end{tabular}
    \caption{
        \textbf{Examples of English-Swahili code-switching in reports along with English translation}.
    }
    \label{tab:lang-examples}
\end{table*}

\begin{figure*}[!t]
    \begin{minipage}{2.25in}
        \includegraphics{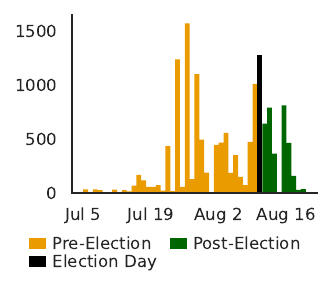}
    \end{minipage}%
    \begin{minipage}{4.0in}
        \centering
        \footnotesize
        \begin{tabular}{lll}
            \toprule
            \colorbox{KenFlagRed}{\color{white} Pre-Election} & \colorbox{black}{\color{white} Election Day} & \colorbox{KenFlagGreen}{\color{white} Post-Election} \\
            \midrule
            PresidentialDebate2022 & RutoThe5THPresident & ElectionsKE2022 \\
            Venezuela & 56.17 & kenyaelections2022 \\
            JKIA & 12,065,803 & President-elect \\
            Smartmatic & KIMS & 138\\
            UNDPKenya & Primary & 0.01 \\
            UchaguziWaAmani & 154 & Serena \\
            LetPeaceWin & Rural & 34C\\
            DCI & Rongai & uploaded \\
            stickers & technicians & announcement \\
            Venezuelans & violating & KenyanElection2022 \\
            \bottomrule
        \end{tabular}
    \end{minipage}%
    \caption{
       \textbf{ Report counts over time} \textit{(left)} \textbf{and words characterizing different phases of the election} \textit{(right)}.
   }
   \label{fig:temporal}
\end{figure*}

\paragraph{Geographic Distribution}
In Figure~\ref{fig:geo-dist} we plot the per capita report count for each county in Kenya.
Population sizes are based on the 2019 Kenyan Census~\cite{Census2019}.
Each report is associated to the county containing its \textit{coordinates} and 
reports using the default location coordinates (\S\ref{sec:dataset-overview}) are omitted.
We observe that the vast majority of the data comes from Nairobi and its surrounding counties.

\paragraph{Temporal Trends}
In Figure~\ref{fig:temporal} we display a histogram of report counts binned by time and colored by election phase (pre-election, election day, post-election), alongside the top 10 tokens characterizing each phase.
The characteristic token lists are determined by identifying the tokens that maximize the probability $P(\textrm{phase} | \textrm{token})$ estimated using a Na\"ive Bayes classifier over word frequencies.

The characteristic tokens provide insight into the key topics of discussion in the different phases of the election.
For instance, the characteristic pre-election tokens largely correspond to a scandal where three Venezuelan employees of the firm Smartmatic were arrested at JKIA airport for possessing stolen election materials~\cite{Venezuelan2022} (this event also corresponds to the spike in reports in late July).
In contrast, the characteristic election day tokens appear to relate to preliminary results and turnout, along with reports of violations, and the characteristic post-election tokens pertain mostly to anticipation of the election results.

\subsection{Data Quality}

We estimate the quality of the annotations provided by the volunteers by sampling 500 reports to be further labeled by an expert annotator (also employed by Ushahidi).
The 500 reports were sampled using the following strategy.
From each of the topics with applicable tags (\textit{Opinions}, \textit{Positive Events}, \textit{Counting and Results}, \textit{Security Issues}, \textit{Voting Issues}, \textit{Polling Station Administration}), we sampled either 100 or 10\% of the reports (whichever was smaller). 
From the remaining topics, we sampled uniformly to reach a total of 500 samples.
This strategy was implemented to take into account the long-tailed nature of the topic distribution (Table \ref{tab:topic-dist}) while ensuring a minimum number of samples for the least frequent topics.
The expert annotator labeled each report with a topic and any applicable tags.

For topic, the results show 50.4\% agreement between the volunteer and expert annotations (252 samples).
This corresponds to a Cohen's kappa of 0.425, indicating moderate reliability~\cite{cohenkappa}.
For tags, considering the set of 252 samples where the expert and the volunteers agreed, the agreement on at least one annotated tag is 71.0\%, reflected by a Cohen's kappa of 0.624.
This indicates good tag reliability once the same topic has been selected. 

Error analysis shows that around half of the samples annotated by the expert with a different topic were assigned to \textit{Opinions} (in particular from \textit{Positive Events} and \textit{Counting and Results}).
This disagreement tends to occur in cases where citizens shared their opinions on events or alleged irregularities rather than reporting factual information.
An example is shown in Figure \ref{fig:inter-annotator-disagreement-example}.
In this case the citizen did not share any factual information on counting irregularities, and therefore this report should be classified as \textit{Opinions} (expert annotation) rather than \textit{Counting and Results} (volunteer annotation).

The overall moderate inter-annotator reliability indicates a degree of subjectivity in the dataset, particularly for the topic label, which we believe is somewhat expected given the dataset size as well as the nature of the annotations provided by the volunteers.
The comparison between the volunteer-provided and expert-provided annotations (used as ``gold standard'') allows us to establish a human baseline for these tasks, with disagreements reflecting the subjectivity according to non-experts.
This subjectivity further supports the need for an automated approach, as this would help improve consistency in the topic and tag classifications.


\begin{figure}[!t]
    \centering
    \footnotesize
    \begin{tabular}{|p{7cm}|}
    \hline
    \textbf{Title:} Confirmation of results by Chebukati as seen by a voter.
     \\
     \textbf{Text:} Chebukati knows when it comes to numbers, @RailaOdinga will win, that's why he's laying ground for nullification of Presidential elections, by messing with the process. Too bad.
     \\ 
     \textbf{Volunteer annotation:} Topic: Counting and Results. Tag: Counting Irregularities.\\
     \textbf{Expert annotation:} Topic: Opinions.\\Tags: Personal Opinion, Negative Opinions.\\
     \hline
    \end{tabular}
    \caption{
    \textbf{Example of disagreement between volunteer and expert annotations}.
    }
    \label{fig:inter-annotator-disagreement-example}
\end{figure}

\section{Methods}

\label{sec:experiments}

We explore the application of \datasetname towards training systems for automated report \textit{categorization} (\S\ref{sec:classification}) and \textit{geotagging} (\S\ref{sec:geotagging}).

\subsection{Automating Categorization}
\label{sec:classification}

Each report is assigned a topic (see Table~\ref{tab:topic-dist}) and optionally annotated with topic-dependent tags (e.g., Table~\ref{tab:tag-example}).
Accordingly, topic prediction is a single-class classification task, while tag prediction is a multi-class classification task.
We train a topic prediction model and six tag prediction models, one for each of the following topics: \textit{Counting and Results}, \textit{Opinions}, \textit{Polling Station Administration}, \textit{Positive Events}, \textit{Security Issues}, \textit{Voting Issues}. The remaining topics lack tags so no tag prediction model is trained.
Additionally, for each task we omit labels with < 20 observations.

For each task we explore two settings: 1) fully supervised learning (FS), and 2) few-shot in-context learning (ICL).
\roberto{Added following details about motivation}
Fully supervised learning allows us to develop categorization models specific to the Uchaguzi platform.
On the other hand, few-shot learning is important to explore from a practical perspective since it allows for quick adaptation of the platform in the case where topics and tags shift across elections and domains.
\rob{I think we could omit the previous two sentences. I get wanting to emphasize the practicality of few-shot learning, but the discussion feels tangential and somewhat distracting here.}
\aoife{I don't find it too distracting here. Another place to put it would be the introduction, but I think it would be more out of place there. I also think it wouldn't be the end of the world to leave it out.}

In the fully supervised setting, we randomly split the volunteer-labeled data into train (80\%) and validation (10\%) sets, and use the expert-labeled data as our test set (which was sub-sampled from the remaining 10\%).
In the few-shot setting, the training examples are sampled from the training split, and the test split is the same one used in the fully supervised setting.
We train independent models for topic prediction and each tag prediction task using the \texttt{transformers} library~\cite{wolf-etal-2020-transformers}.
We fine-tune \texttt{XLM-RoBERTa-base}~\cite{conneau-etal-2020-unsupervised} for 100 epochs and select the best checkpoint using validation set performance (hyperparameters reported in Appendix~\ref{sec:appendix-classification-experiments}).

In the few-shot setting, we perform 25-shot learning with \texttt{mixtral-8x7b}~\cite{jiang2024mixtral}, \texttt{llama-3.1-70b}~\cite{dubey2024llama3herdmodels}, and \texttt{gpt-4o}~\cite{openai2024gpt4}. The LLMs chosen represent those widely used at the time of experimentation, and based on the available analyses, have some Swahili proficiency in the few-shot setting~\cite{Ochieng2024,openai2024gpt4}. \neema{Rephrased the motivation behind choice of LLMs, included refs.}
We experiment with varying the number of examples used for in-context learning as well as with including topic and tag descriptions in the prompts.
The final prompts are selected based on performance on the validation set and are shown in Appendix~\ref{sec:appendix-classification-experiments}.


\subsection{Automating Geotagging}
\label{sec:geotagging}

We decompose geotagging into two sub-tasks: 1) \textit{location extraction}, extracting mentioned locations from reports, and 2) \textit{geocoding}, retrieving their corresponding point coordinates.

\begin{figure}[!t]
    \centering
    \small
    \begin{tabular}{|p{7cm}|}
    \hline
     \textbf{Title:} IEBC Officials Deliver Results In \textcolor{KenFlagGreen}{\textbf{Saboti}} Constituency\\
     \textbf{Text:} IEBC officials deliver results from various polling stations in \textcolor{KenFlagGreen}{\textbf{Saboti}} constituency to tallying centre at \textcolor{KenFlagGreen}{\textbf{St Joseph's Girls High school}}.
\#KenyaDecides2022\\
    \textbf{Location:} Saboti, St Joseph's Girls High school\\
    \hline 
    \end{tabular}
    \caption{
    \textbf{Location extraction}. 
    We retrieve all location mentions (\textcolor{KenFlagGreen}{\textbf{green}}) in the title or text of a report.}
    \label{fig:location-extraction-examples}
\end{figure}

\paragraph{Location Extraction} 

Given a report title and text, we detect all mentioned locations (Figure \ref{fig:location-extraction-examples}).
We explore NER and few-shot ICL approaches for this sub-task.
\roberto{Added following details about motivation}
As is the case for the categorization tasks, the decision to use ICL for location extraction is motivated by the adaptability of this LLM-based approach, along with the fact that the parametric knowledge of these LLMs encapsulates many Kenyan locations.
Additionally, we consider only reports for which we could manually verify that the annotated location appears in their title or text (147 samples).
Due to this size, fine-tuning a location extraction model was unfeasible.
We experiment with \texttt{wikiNEuRal}~\cite{tedeschi-etal-2021-wikineural-combined} for NER, while the ICL approach relies on the same models as in \S\ref{sec:classification}.

\paragraph{Geocoding} 

The extracted location names are used to query corresponding point coordinates from OpenStreetMap (OSM) using the Nominatim API\footnote{\url{https://nominatim.org/}}.
We evaluate geocoding on a total of 2,818 samples: the 147 samples used for the location extraction evaluation,
and, additionally, a silver dataset of 2,681 reports that do not have annotated locations but have associated coordinates and likely include a location mention.
This latter set allows us to understand the extent to which location extraction systems can recover location names for reports with incomplete annotations.
To obtain these evaluation sets, we begin with all reports with coordinates that do not correspond to a \textit{default location} (\S\ref{sec:dataset-overview}).
We use zero-shot prompting with \texttt{gpt-4o} (Appendix \ref{sec:appendix-geotagging-prompts}) to identify and filter out reports that do not mention locations.
We choose \texttt{gpt-4o} as it has the highest recall for location extraction (\S\ref{sec:location-extraction-results}).
We employ this approach because we observe a number of reports with labeled location but no location mentions in their title or text, making them unsuitable for geocoding.

\section{Evaluation}

\subsection{Automating Categorization}
\label{sec:categorization-results}

Performance on the topic prediction and tag prediction tasks is measured on the test set using micro-averaged and macro-averaged $F_1$ scores.

\begin{table}[!t]
    \centering
    \small
    \begin{tabular}{lcc}
    \toprule
    \textbf{Model} & \textbf{Micro $F_1$} & \textbf{Macro $F_1$}\\
      \midrule
      Human annotation (baseline)  & 50.4 & 48.6 \\
      \midrule
      \texttt{XLM-RoBERTa-base} (FS)  & 49.4 & 27.9 \\
      \texttt{gpt-4o} (ICL)  & 47.8 & 36.2 \\
      \texttt{mixtral-8x7b} (ICL) & 45.0 & 29.3\\
      \texttt{llama-3.1-70b} (ICL) & \textbf{54.6} & \textbf{39.7}\\
    \bottomrule
    \end{tabular}
    \caption{\textbf{Topic prediction results.}}
    \label{tab:topic-prediction-results-annotation-dataset}
\end{table}

\subsubsection{Topic Prediction} 
\label{sec:topic-prediction-results}
The results for all models as well as the volunteer annotations (human baseline) are presented in Table \ref{tab:topic-prediction-results-annotation-dataset}.
We observe that, overall, the ICL models are competitive with the FS model, with \texttt{llama-3.1-70b} outperforming all other models across both metrics.
Table \ref{tab:topic-predictions-results-llama3-xlm-roberta} breaks down the $F_1$ score by class for \texttt{XLM-RoBERTa} (FS) and \texttt{llama-3.1-70b} (ICL).
While the FS model performs similarly on classes with many training examples (hence the comparable micro-averaged $F_1$), the ICL model outperforms on classes with fewer training examples (e.g., \textit{Irrelevant Report} and \textit{Political Rallies}), thus leading to a higher macro-averaged $F_1$.
Both models fail to predict the \textit{Polling Station Administration} and \textit{Staffing Issues} topics.
This is due to the fact that these topics have few training examples in the dataset (Table~\ref{tab:topic-dist}) and are hard to distinguish from semantically related topics, such as \textit{Security Issues} and \textit{Voting Issues}.

\begin{table}[!t]
    \centering
    \setlength{\tabcolsep}{0.1pt}
    \small
    \begin{tabular}{lcc}
    \toprule
    \textbf{Topic} & \texttt{XLM-RoBERTa} (FS) & \texttt{ llama-3.1-70b} (ICL) \\ 
      \midrule
      Opinions & 63.7 & \textbf{69.4} \\ 
      Media Reports & 51.1 & \textbf{51.6} \\ 
      Count. and Res. & 50.6 & \textbf{58.0} \\ 
      Positive Events & 43.7 & \textbf{44.0} \\ 
      Security Issues & 41.7 & \textbf{57.1} \\ 
      Voting Issues & \textbf{28.6} & 26.7 \\ 
      Irrelevant Report & 0.0 & \textbf{64.0} \\ 
      Political Rallies & 0.0 & \textbf{26.7} \\ 
      Polling Admin. & 0.0 & 0.0 \\ 
      Staffing Issues & 0.0 & 0.0 \\ 
    \bottomrule
    \end{tabular}
    \caption{\textbf{$F_1$ breakdown for the} \texttt{XLM-RoBERTa} (FS) \textbf{and} \texttt{llama-3.1-70b} (ICL) \textbf{topic prediction models.}}
    \label{tab:topic-predictions-results-llama3-xlm-roberta}
\end{table}

\begin{figure}[!t]
    \centering
    \footnotesize
    \begin{tabular}{|p{7cm}|}
    \hline
     \textbf{Title:} Life beyond elections
     \\ 
     \textbf{Text:} After elections we want to move with our life as if nothing happened.\\
     \textbf{Topic:} Opinions\\
     \hline
    \end{tabular}

    \vspace{.5em}
    
    \begin{tabular}{ll}     
     \texttt{XLM-RoBERTa-base}: & Positive Events\\
     \texttt{gpt-4o}: & Opinions\\
     \texttt{mixtral-8x7b}: & Positive Events\\
     \texttt{llama-3.1-70b}: & Positive Events\\
    \end{tabular}

    \vspace{1em}

    \begin{tabular}{|p{7cm}|}
    \hline
    \textbf{Title:} No voting going on at Hospital Hill Primary
     \\
     \textbf{Text:} @IEBCKenya 6:34AM No voting going on at Hospital Hill Primary. No voting material \& no IEBC officials.
     \\ 
     \textbf{Topic:} Polling Station Administration\\
     \hline
    \end{tabular}

    \vspace{.5em}
    
    \begin{tabular}{ll}
    \texttt{XLM-RoBERTa-base}: & Security Issues\\
    \texttt{gpt-4o}: & Voting Issues\\
    \texttt{mixtral-8x7b}: & Voting Issues\\
    \texttt{llama-3.1-70b}: & Voting Issues\\
    \end{tabular}
     
    
    \caption{
    \textbf{Examples of inaccurate topic prediction}.
    }
    \label{fig:topic-prediction-error-analysis-examples}
\end{figure}

In Figure \ref{fig:topic-prediction-error-analysis-examples} we present two examples of incorrect model predictions.
The first example represents the most common type of mistake across all models. Most models focus on the positive nature of this message, and do not recognize that it does not relate to any factual event (and therefore should be classified as \textit{Opinions}).
In the second example, most models identify that the citizen is reporting on issues pertaining to the voting process, but fail to understand that the report is about the operations of a polling station and therefore should be classified as \textit{Polling Station Administration}.
This type of mistake is more subtle and typically involves related topics (such as \textit{Polling Station Administration}, \textit{Voting Issues}, and \textit{Security Issues}).

Finally, the results in Table \ref{tab:topic-prediction-results-annotation-dataset} also show that the performance of \texttt{llama-3.1-70b} is competitive with the human baseline, with lower macro-averaged $F_1$ but higher overall accuracy.
This suggests that this model could be used to create silver annotations on the unlabeled data, and aid volunteers in assigning topics. 

\begin{table*}[!t]
    \centering
    \setlength{\tabcolsep}{2pt}
    \small
    \begin{tabular}{lcccc}
        \toprule
        \textbf{Task}           & \texttt{XLM-RoBERTa} (FS) & \texttt{gpt-4o} (ICL) & \texttt{mixtral-8x7b} (ICL) & \texttt{llama-3.1-70b} (ICL) \\
        \midrule
        Counting and Res.       &         38.4  &         49.5  &         44.9  & \textbf{52.3} \\
        Opinions                & \textbf{51.0} &         49.5  &         35.4  &         38.2  \\
        Polling Admin.          &         77.8  &         76.2  & \textbf{87.0} &         81.8  \\
        Positive Events         &         55.4  & \textbf{65.7} &         56.7  &         59.2  \\
        Security Issues         &         58.3  &         64.3  &         73.3  & \textbf{74.1}  \\
        Voting Issues           & \textbf{60.8} &         38.7  &         48.3  &         38.7  \\
        \bottomrule
    \end{tabular}
    \caption{\textbf{Tag prediction results.} Performance is measured using micro-averaged $F_1$.}
    \label{tab:tag-prediction-results}
\end{table*}

\subsubsection{Tag Prediction} 
\label{sec:tag-prediction-results}

Table~\ref{tab:tag-prediction-results} shows the tag prediction results.
Overall, we observe fairly comparable performance for the different models.
When larger training sets are available (e.g., \textit{Opinions}), the FS model outperforms ICL models. 
In contrast, few-shot ICL performs better on the tasks with fewer training examples, with \texttt{llama-3.1-70b} as the best performing model on two tasks (\textit{Counting and Results}, \textit{Security Issues}). 
Despite the limited dataset size, all models achieve quite good performance on the \textit{Polling Station Administration} task, with \texttt{mixtral-8x7b} obtaining a micro-averaged $F_1$ score of 87.0.

In Figure \ref{fig:tag-prediction-error-analysis-examples} we show two examples of incorrect model predictions.
In the first example (belonging to \textit{Voting Issues}), while the FS model does not predict any tags, the ICL models correctly predict that the report pertains to voting irregularities. However, they fail to recognize the more subtle nature of the issue, related to voting kits rather than registration procedures.
In the second example (belonging to \textit{Counting and Results}), only \texttt{llama-3.1-70b} is able to identify that the text implies counting irregularities.
These mistakes occur when models cannot discriminate between semantically-related tags and represent the most common type of mistakes.

\begin{figure}[!t]
    \centering
    \footnotesize
    \begin{tabular}{|p{7cm}|}
    \hline
     \textbf{Title:} KIEMS Kit Failure
     \\ 
     \textbf{Text:} Limbine Primary School polling centre has 3 polling stations; and polling station 3 has been closed from 1:00pm due to kit failure. What are the voters going to do by 6:00pm? IEBC should do something.\\
     \textbf{Topic:} Voting Issues\\
     \textbf{Tag:} Voter Integrity Irregularities\\
     \hline
    \end{tabular}

    \vspace{0.5em}
    
    \begin{tabular}{ll}     
     \texttt{XLM-RoBERTa-base}: & $-$\\
     \texttt{gpt-4o}: & Voter Registration Irregularities\\
     \texttt{mixtral-8x7b}: & Voter Assistance Irregularities\\
     \texttt{llama-3.1-70b}: & Voter Registration Irregularities\\
    \end{tabular}
    
    \vspace{1em}
    
    \begin{tabular}{|p{7cm}|}
    \hline
     \textbf{Title:} Confusion over IEBC results and projections
     \\ 
     \textbf{Text:} Forms delivered are over 88\% but the information being translated in votes is still less than 20\%.\\
     \textbf{Topic:} Counting and Results\\
     \textbf{Tag:} Counting Irregularities\\
     \hline
    \end{tabular}

    \vspace{0.5em}
    
    \begin{tabular}{ll}     
     \texttt{XLM-RoBERTa-base}: & Provisional Citizen Results\\
     \texttt{gpt-4o}: & Failure to announce results by\\
                            & IEBC officials\\
     \texttt{mixtral-8x7b}: & Failure to announce results by\\
                            & IEBC officials\\
     \texttt{llama-3.1-70b}: & Counting Irregularities\\
    \end{tabular}
    
    \caption{
    \textbf{Examples of inaccurate tag prediction}. Minor edits to the original texts for readability.
    }
    \label{fig:tag-prediction-error-analysis-examples}
\end{figure}

\subsection{Automated Geotagging}
\label{sec:geotagging-results}

\subsubsection{Location Extraction} 
\label{sec:location-extraction-results}

\begin{table}[!t]
    \centering
    \setlength{\tabcolsep}{3pt}
    \footnotesize
    \begin{tabular}{lp{1.5cm}p{1.5cm}p{1.5cm}}
    \toprule
    \textbf{Model} & \textbf{\% Empty Pred. ($\downarrow$)} & \textbf{Exact Match ($\uparrow$)} & \textbf{ROUGE-L} ($\uparrow$)\\
      \midrule
      \texttt{wikiNEuRal} & 20.4 & 1.4  & 14.2 \\
       \texttt{gpt-4o} & \textbf{3.4} & 46.3 & 71.5 \\
       \texttt{mixtral-8x7b} & 10.2 & 40.1 &  67.6 \\
       \texttt{llama-3.1-70b} & 4.8 & \textbf{49.0} & \textbf{75.3} \\
    \bottomrule
    \end{tabular}
    \caption{\textbf{Location extraction results for 147 samples with labeled location.}
    Exact match and ROUGE-L are averaged across extracted locations.}
    \label{tab:toponym-extraction-results}
\end{table}

Location extraction is evaluated along two dimensions: 1) the percentage of samples for which no location is extracted, and 2) how similar the extracted locations are to the annotations using text similarity metrics, such as 
exact match and ROUGE-L\footnote{ROUGE-L is chosen to fairly evaluate generative approaches, as they may produce location names that resemble but do not exactly match spans in the source text.} \cite{lin-2004-rouge}. 
Table~\ref{tab:toponym-extraction-results} shows larger models achieve the highest performance in both exact match and ROUGE-L, with
\texttt{llama-3.1-70b} outperforming the other models.
We observe a large gap in performance between the mBERT-based \texttt{wikiNEuRal} model and the ICL approaches.
Although \texttt{wikiNEuRal} is generally able to identify mentions of large cities (e.g., Nairobi, Mombasa), it performs poorly at detecting specific landmarks, hence the overall very low performance.
As \texttt{WikiNEuRal} was trained on 50k English Wikipedia articles, it is unlikely that the training data covered the specific locations present in our evaluation set.

\subsubsection{Geocoding} 
\label{sec:geocoding-results}

We report three metrics commonly used to evaluate geocoding: coverage (percentage of queries that return coordinates), accuracy at 161 km (Acc@161km), and area under the curve (AUC).
Acc@161km computes the percentage of locations which are correctly predicted to within 161 km (100 miles) of the actual location.
On the other hand, AUC, computed using the trapezoidal rule~\cite{yeh2002using}, measures the error across all instances while minimizing the importance of outliers. 
AUC for geocoding was introduced by~\citet{jurgens-auc-2015} and it allows for relative comparison of geocoders with similar Acc@161km scores.

We present results in Table~\ref{tab:geo-coding-results-loc} (samples with labeled location) and Table~\ref{tab:geo-coding-results-loc-less} (samples without labeled location). Where no coordinates are returned, we use Null Island\footnote{\url{https://en.wikipedia.org/wiki/Null_Island}} as the reference to compute the error.
For the oracle in Table~\ref{tab:geo-coding-results-loc}, the annotated location names serve as input queries for geocoding.

\begin{table}[t]
    \setlength{\tabcolsep}{4pt}
    \centering
    \small
    \begin{tabular}{lp{1.1cm}p{1.4cm}p{1.1cm}}
    \toprule
     \textbf{Model} & \textbf{Cov.} ($\uparrow$) &\textbf{Acc@ {161km}} ($\uparrow$) & \textbf{AUC} ($\downarrow$)\\
     \midrule
      \texttt{oracle} & 53.7  & 43.5 & 337\\
      \midrule
      \texttt{wikiNEuRal} & 79.6 & 6.1 & 495  \\
      \texttt{gpt-4o} (ICL) & 86.4 & 66.7 & 156\\
      \texttt{mixtral-8x7b} (ICL) & 78.2  &  62.6  & 209 \\
      \texttt{llama-3.1-70b} (ICL) & \textbf{87.1} & \textbf{68.7} & \textbf{123} \\
     \bottomrule
    \end{tabular}
    \caption{
    \textbf{Geocoding results for 147 samples with labeled location.}
    AUC is expressed in 1,000s of km$^2$.
    }
    \label{tab:geo-coding-results-loc}
\end{table}

\begin{table}[t]
    \setlength{\tabcolsep}{4pt}
    \centering
    \small
    \begin{tabular}{lp{1.1cm}p{1.4cm}p{1.1cm}}
    \toprule
     \textbf{Model} & \textbf{Cov.} ($\uparrow$) &\textbf{Acc@ {161km}} ($\uparrow$) & \textbf{AUC} ($\downarrow$)\\
      \midrule
      \texttt{wikiNEuRal} & 36.3 & 1.9 &  9,218\\
      \texttt{gpt-4o} (ICL) & \textbf{80.6} & 38.2 & 4,245\\
      \texttt{mixtral-8x7b} (ICL) & 63.6 & \textbf{41.1}  & 5,540\\
      \texttt{llama-3.1-70b} (ICL) & 80.2 & 38.2 & \textbf{3,674} \\
     \bottomrule
    \end{tabular}
    \caption{
    \textbf{Geocoding results for 2,681 samples without labeled location.}
    AUC is expressed in 1,000s of km$^2$.
    }
    \label{tab:geo-coding-results-loc-less}
\end{table}

For samples with labeled location (Table~\ref{tab:geo-coding-results-loc}), we observe that while all location extraction models have good coverage, the ICL approaches clearly outperform the NER approach on accuracy and AUC.
In particular, the higher location extraction performance observed in Table~\ref{tab:toponym-extraction-results} for \texttt{llama3.1-70b} provides this model with an advantage over all other approaches for the downstream task.
Across approaches, geocoding errors most frequently tend to occur for specific building names.
The oracle performs poorly compared to the ICL approaches because the annotated locations can be more specific and therefore less likely to be resolved by the OSM query than the less precise locations extracted by the ICL models.
Finally, we observe that although the locations extracted using \texttt{wikiNEuRal} achieve high coverage, the returned coordinates are mostly inaccurate.
 
The results for samples without labeled location (Table~\ref{tab:geo-coding-results-loc-less}) show comparable performance among the ICL models.
Comparing the results for samples with labeled locations
and those without,
we observe a marked decrease in performance across all metrics, most starkly for Acc@161km and AUC.
However, the results show that the ICL methods are still able to correctly geocode around 40\% of reports without explicit location annotations, providing evidence that a lack of annotated location does not necessarily mean that the report does not mention any.
Our geocoding approaches tend to fail for two reasons: either the queried location is not present in the OSM database, or the location is resolved but OSM returns coordinates that are vastly different from those in the labeled data, resulting in a significant error.
We believe better map coverage, particularly for building names and landmarks, could further improve performance.

\begin{figure}[!tb]
    \centering
    \small
    \begin{tabular}{|p{7cm}|}
    \hline
    \textbf{Title:} Civic Education Campaign\\
    \textbf{Text:} Day 2 of our Civic Education campaign  we are in \textcolor{KenFlagGreen}{\textbf{Changamwe Social Hall}} having a dialogue with young people on why the should vote and the do's and dont's during the voting day with @siasaplace and @IEBC\_YCC \#GE2022 \\
    \textbf{Coordinates:}  $(-4.02084, 39.62748)$  \\
    \hline
    \end{tabular}

    \vspace{0.2em}

    \begin{tabular}{ll}
    \texttt{gpt-4o}: & Changamwe Social Hall\\
    Retrieved Coordinates: & $-$\\
    \end{tabular}
    
    
    \caption{
        \textbf{Example of failed geocoding}.
    }
    \label{fig:geotagging-predictions-example}
\end{figure}

Figure~\ref{fig:geotagging-predictions-example} illustrates such an example.
In this case the report was geotagged with coordinates but no explicit location was added.
While \texttt{gpt-4o} is able to extract the correct location mention, the exact building name cannot be resolved into point coordinates using OpenStreetMap and therefore \texttt{gpt-4o} fails to geocode the report.

\section{Related Work}



Alongside the rise of social media, election analysis has become a popular topic in NLP. 
Studies in this field have typically focused on tasks such as electoral sentiment analysis~\cite{wang_system_2012,schmidt_sentiment_2022,hellwig_transformer-based_2023}, predicting election outcomes \cite{bermingham_using_2011,tjong_kim_sang_predicting_2012,sanders_optimising_2020}, detecting online political communities \cite{wan_ranking_2015,abdine_political_2022}, electioneering \cite{baran_electoral_2022}, predicting success rates of campaigning strategies \cite{litvak_detection_2022,mohapatra_sentiment_2022,bhaumik_adapting_2023}, and analyzing election fraud claims~\cite{Abilov_Hua_Matatov_Amir_Naaman_2021}.
Our work differs from these studies in a number of ways.

Firstly, existing literature mostly centers on elections in the United States or Europe, whereas \datasetname covers Kenyan elections and politics.
While there are other works that focus on the 2022 elections \cite{amol-etal-2024-politikweli}, their emphasis is on creating a political misinformation dataset.
Secondly, \datasetname is novel in its focus on detecting and categorizing reports of electoral misconduct, voting issues, and related acts of violence.
Although there exist other datasets studying issues of civil unrest,
\citet{chinta-etal-2021-study} does not include fine-grained topic and geographical information, while the datasets introduced in \citet{acled-2010} and \citet{sundberg-2013-ucdp} are based on news articles and structured reports from NGOs rather than citizen reports. 
\datasetname is unique among these datasets as it also links reported events to their locations.
Lastly, in addition to being a useful resource for training models to identify reports of electoral misconduct, we believe \datasetname could also be of value to the topic of geotagging noisy social media posts~\cite{seeberger-riedhammer-2022-enhancing,li_leveraging_2024}, as the majority of reports are X (formerly Twitter) posts.


\section{Conclusion}

In this paper we introduced \datasetname, a dataset of 14k citizens' opinions and reports on the 2022 Kenyan General Election, including mentions of electoral misconduct and election-related acts of violence.
We demonstrated potential applications of this dataset by training and benchmarking language models for automated report categorization and geotagging, achieving viable performance while leaving ample room for further improvement.
Initial explorations integrating these models into a live system indicate that the performance of the topic and tag prediction models can improve efficiency in generating insights in rapidly evolving situations \cite{ushahidi-report-finance-bill-protests-2024},
while next steps are needed to evaluate the geotagging models in this setting. 
We additionally presented preliminary analyses of the presence of different languages, code-switching, and geo-spatial and temporal properties of the reports in this dataset.
We believe these demonstrate the dataset's potential for deeper sociolinguistic analyses.
Finally, we hope this dataset will be a valuable resource for studying NLP applications to election integrity monitoring in areas of the world where such systems could be crucial to help ensure fair elections, thus further advancing AI for Social Good efforts.

\section{Limitations}

The topic labels used to annotate the dataset are not specific to the context of Kenyan elections, and therefore we believe that our approach for the topic prediction task can be applied to elections in other geographical regions.
On the other hand, the topic-dependent tags are specific to Kenya (e.g., they reference the Independent Electoral and Boundaries Commission).
Tasks focusing on elections in other countries would require a new set of applicable tags to be defined.

We acknowledge that for some tag prediction tasks test set sizes are limited, restricting our ability to compare the performance of different models. We believe this further highlights the need for datasets such as \datasetname, as manually processing large amount of reports to collect a sufficient number of examples for each topic and tag category is intractable.

\section{Ethical Considerations}

We would like to emphasize that the dataset presented in this work contains only data that is already publicly available online and collected in accordance with the terms of service of the Uchaguzi platform.
None of the reports are confidential, and care has been taken to remove any personally identifiable information (such as names and phone numbers).
Furthermore, the geographic coordinates associated to the reports reflect the physical location of the reported event and \emph{not} the location of the citizen reporting the event.

In order to access and download the full dataset, researchers will need to fill out a licensing agreement prohibiting them from using the data for malicious purposes.
Accordingly, we do not anticipate the publication of this dataset would endanger any person or persons.

\section*{Acknowledgments}


This work was made possible by Dataminr's AI for Good program and we are grateful to Sirene Abou-Chakra and Jessie End for their support during this project.
We thank the partners at Ushahidi for their support and collaboration.
We also wish to thank the reviewers and meta-reviewer for their valuable feedback.

\bibliography{rebiber_custom}

\clearpage
\appendix

\counterwithin{figure}{section}
\counterwithin{table}{section}
\renewcommand\thefigure{\thesection\arabic{figure}}
\renewcommand\thetable{\thesection\arabic{table}}

\section{Tag Sets}
\label{sec:appendix-tag-assignment-categories}

We provide the complete list of applicable tags within each topic in Tables \ref{tab:counting_and_results_table}-\ref{tab:voting_issues_table} along with their prevalence.
Reports belonging to the following topics are not further assigned any tags: \textit{Media Reports}, \textit{Political Rallies}, and \textit{Irrelevant Report}.
Table \ref{tab:topic-examples} shows the list of topics with example messages.

\begin{table}[!th]
    \small
    \centering
    \setlength{\tabcolsep}{1pt}
    \begin{tabular}{lr}
        \toprule
        \textbf{Tag}  &  \textbf{Count} \\
        \midrule
        Unofficial proclamation of results & 220 \\
        Provisional citizen results & 173 \\
        Counting irregularities & 74 \\
        Official IEBC results & 64 \\
        Protest over declared results & 43 \\
        Failure to announce results by IEBC official & 33 \\
        Fake 34 result forms & 17 \\
        Party agent irregularities & 5 \\
        Irregularities with transportation of ballot boxes & 3 \\
        \bottomrule
    \end{tabular}
    \caption{\textbf{Counting and Results}.}
    \label{tab:counting_and_results_table}
\end{table}

\begin{table}[!th]
    \small
    \centering
    \begin{tabular}{lr}
        \toprule
        \textbf{Tag}  &  \textbf{Count} \\
        \midrule
        Personal opinion & 3,577 \\
        Positive opinions & 2,392 \\
        Negative opinions & 1,887 \\
        Neutral & 1,543 \\
        Peace messages & 435 \\
        \bottomrule
    \end{tabular}
    \caption{\textbf{Opinions}.}
    \label{tab:opinions_table}
\end{table}

\begin{table}[!th]
    \small
    \centering
    \begin{tabular}{lr}
        \toprule
        \textbf{Tag}  &  \textbf{Count} \\
        \midrule
        Biometric voter registration (BVR) issues & 38 \\
        Polling station logistical issues & 38 \\
        Low voter turn out & 23 \\
        Missing/inadequate voting materials & 14 \\
        Ballot box irregularities & 9 \\
        Polling station closed before voting concluded & 5 \\
        Campaign material in polling station & 3 \\
        High voter turn out & 2 \\
        \bottomrule
    \end{tabular}
    \caption{\textbf{Polling Station Administration}.}
    \label{tab:psa_table}
\end{table}

\begin{table}[!th]
    \small
    \centering
    \begin{tabular}{lr}
        \toprule
        \textbf{Tag}  &  \textbf{Count} \\
        \midrule
        Citizen led initiatives to promote peace & 925 \\
        Everything fine & 306 \\
        Organization led initiatives & 58 \\
        Police peace efforts & 18 \\
        Civic education & 5 \\
        \bottomrule
    \end{tabular}
    \caption{\textbf{Positive Events}.}
    \label{tab:positive_events_table}
\end{table}

\begin{table}[!th]
    \small
    \centering
    \begin{tabular}{lr}
        \toprule
        \textbf{Tag}  &  \textbf{Count} \\
        \midrule
        Rumors & 117 \\
        Dangerous speech & 62 \\
        Violent attacks & 32 \\
        Abductions/kidnapping & 26 \\
        Mobilisation towards violence & 26 \\
        Demonstrations & 24 \\
        Heavy police presence & 22 \\
        Vandalism and physical attacks on property & 21 \\
        Riots & 15 \\
        Presence of weapons & 10 \\
        Armed clashes & 8 \\
        Ambush & 7 \\
        Eviction/population displacement & 4 \\
        Police brutality & 3 \\
        Sexual and gender based violence & 3 \\
        \bottomrule
    \end{tabular}
    \caption{\textbf{Security Issues}.}
    \label{tab:security_issues_table}
\end{table}

\begin{table}[!th]
    \small
    \centering
    \begin{tabular}{p{5.0cm}r}
        \toprule
        \textbf{Tag}  &  \textbf{Count} \\
        \midrule
        IEBC officials not acting in accordance to set rules & 65 \\
        Absence or insufficient number of IEBC officials/staff at polling station opening & 8 \\
        Observers/media blocked from entering polling station & 2 \\
        Absence or insufficient number of law enforcement officials at polling station & 1 \\
        \bottomrule
    \end{tabular}
    \caption{\textbf{Staffing Issues}.}
    \label{tab:staffing_issues_table}
\end{table}

\begin{table}[!th]
    \small
    \centering
    \begin{tabular}{lr}
        \toprule
        \textbf{Tag}  &  \textbf{Count} \\
        \midrule
        Voting irregularities & 69 \\
        Voter integrity irregularities & 66 \\
        Voter registration irregularities & 34 \\
        Civic education gap & 33 \\
        Voter assistance irregularities & 28 \\
        Alleged rigging & 21 \\
        Voters issued invalid ballot papers & 11 \\
        Voting suspended/postponed & 10 \\
        Purchasing of voters cards & 9 \\
        Voter intimidation and blockage & 7 \\
        \bottomrule
    \end{tabular}
    \caption{\textbf{Voting Issues}.}
    \label{tab:voting_issues_table}
\end{table}

\begin{table*}[t]
    \centering
    \small
    \begin{tabular}{ll}
       \toprule
       \textbf{Topic}  &  \textbf{Example Message} \\
       \midrule
       Opinions & So shameful that \#wafulachebukati IEBC Chairman is once again in headlines for\\ & the same wrong reasons of 2017. In civilized countries once you have been found\\ & to lack integrity one resigns.\vspace{0.1cm} \\
       Media Reports & William Ruto: "We made a commitment we are going to have a Diaspora Ministry\\ & so that the many issues that concern our people in the diaspora are sorted out"\\ & \#KenyaDecides\vspace{0.1cm} \\ 
       Positive Events & There was a peaceful and organised elections in bondo constituency siaya county\vspace{0.1cm} \\ 
       Counting and Results & William Ruto's losses 10,000 votes in Kiambu constituency as  IEBC corrects the\\ & votes from 51,050 to 41,050.\vspace{0.1cm} \\ 
       Security Issues & Reports of road blockage on the outer ring road in NBO. Southern bypass near\\ & Kibra also seeing worsening disruption. Avoid area.\vspace{0.1cm} \\ 
       Voting Issues & Voters funding it difficult to get their names in voters register; some of those who\\ & talked to us say IEBC officials can't help them\vspace{0.1cm} \\ 
       Political Rallies & Deputy President (incoming) Rigathi Gachagua will tour Narok \& Kajiado counties\\ & today. Kenya Kwanza RG's Itinerary: 1. Narok Town 2. Nairagie Enkare 3. Uwaso\\ & 4. Ilasit (KAJIADO South) 5. Loitoktok 6. Kimana IEBC Dismus\vspace{0.1cm} \\ 
       Polling Station Administration & IEBC failed to deliver manual registers to every polling station. Just seen 50 votes\\ & for Baba getting lost in my station\vspace{0.1cm} \\ 
       Staffing Issues & Eight IEBC officials from Homabay, Kisumu and Bungoma counties have been\\ & arrested and sacked after they were found meeting two candidates for parliamentary\\ & and county assembly.\vspace{0.1cm} \\ 
       Irrelevant Report & If i dont have ID i can vote? \\ 
       \bottomrule
    \end{tabular}
    \caption{
        \textbf{List of topics and example messages}.
    }
    \label{tab:topic-examples}
\end{table*}

\clearpage

\section{Classification}
\label{sec:appendix-classification-experiments}

For the fully-supervised setting, we finetune models using AdamW~\cite{loshchilov2018decoupled} optimizer with the hyperparameters specified in Table~\ref{tab:classification-hyperparams}.
Finetuning was performed on machines with one NVIDIA A10g GPU (24 GB of VRAM).

In the few-shot in-context learning setting, the prompt for the topic prediction task is shown in Table~\ref{tab:prompt-topic-prediction}, while the prompts for the tag prediction tasks are shown in Tables~\ref{tab:prompt-counting-and-results}-\ref{tab:prompt-voting-issues}.
For tag prediction, outputs are requested to be provided as JSON-formatted arrays, and are parsed by truncating the output string after the first occurrence of a {\tt ]} character and then passed through a JSON parser.
We did not encounter formatting issues.


\begin{table}[!th]
    \centering
    \begin{tabular}{ll}
        \hline
         Learning Rate & 1e-5 \\
         $\beta_1$ & 0.9 \\
         $\beta_2$ & 0.999 \\
         $\epsilon$ & 1e-8 \\ 
         Max. Grad. Norm & 1.0 \\
         Batch Size & 8 \\ 
         \hline
    \end{tabular}
    \caption{
        {\bf Fine-tuning hyperparameters.}
    }
    \label{tab:classification-hyperparams}
\end{table}

\begin{table}[!th]
    \centering
    \small
    \begin{tabular}{|p{7cm}|}
    \hline
     \texttt{You are a topic classifier. Given an input text you will classify it with one and only one of the following topics:}
     
     \texttt{Counting and Results}
     
     \texttt{Staffing Issues}

     \texttt{Positive Events}

     \texttt{Opinions}
     
     \texttt{Political Rallies}
     
     \texttt{Polling Station Administration}
     
     \texttt{Irrelevant Report}
     
     \texttt{Security Issues}
     
     \texttt{Media Reports}
     
     \texttt{Voting Issues}
     \\
    \hline
    \end{tabular}
    \caption{{\bf Prompt for topic prediction.}}
    \label{tab:prompt-topic-prediction}
\end{table}

\begin{table}[!th]
    \centering
    \small
    \tt
    \begin{tabular}{|p{7cm}|}
    \hline
    You are a topic classifier. Given an input text you will classify it with zero or more of the following topics output in a JSON array format: \\
    Label: Failure to announce results by IEBC officials, Description: Failure to announce results by IEBC officials \\
    Label: Protest over declared results, Description: Violence and demonstrations revolving around voting results \\
    Label: Unofficial proclamation of results, Description: Unofficial proclamation of results \\
    Label: Official IEBC results, Description: Official IEBC results \\ 
    Label: Provisional Citizen Results, Description: Provisional Citizen Results \\
    Label: Counting Irregularities, Description: Involves Ballot Papers not Being Counted in a Transparent Manner, Observers or Party Agents not Allowed In The Hall During Vote Counting, Spoiled Ballot Papers not Properly Preserved For Review, Intimidation of Counting Officials \& Observers, Error or Omission In Computing or Completing Tally Sheets, Unusually Many Rejected/Spoilt Ballot Papers, Officials Tallying Wrong/Tampered Results, Officials not Reporting Results At Prescribed Time, etc.
     \\
    \hline
    \end{tabular}
    \caption{{\bf Prompt for \emph{Counting and Results} tag prediction.}}
    \label{tab:prompt-counting-and-results}
\end{table}

\begin{table}[!th]
    \centering
    \small
    \tt
    \begin{tabular}{|p{7cm}|}
    \hline
    You are a topic classifier. Given an input text you will classify it with zero or more of the following topics output in a JSON array format: \\
    Negative opinions \\
    Neutral \\
    Peace messages \\
    Personal Opinion \\
    Positive Opinions \\
    \hline
    \end{tabular}
    \caption{{\bf Prompt for \emph{Opinions} tag prediction.}}
    \label{tab:prompt-opinions}
\end{table}

\begin{table}[!th]
    \centering
    \small
    \tt
    \begin{tabular}{|p{7cm}|}
    \hline
    You are a topic classifier. Given an input text you will classify it with zero or more of the following topics output in a JSON array format: \\
    BVR issues \\
    Low voter turn out \\
    Polling station logistical issues \\
    \hline
    \end{tabular}
    \caption{{\bf Prompt for \emph{Polling Station Administration} tag prediction.}}
    \label{tab:prompt-psa}
\end{table}

\begin{table}[!th]
    \centering
    \small
    \tt
    \begin{tabular}{|p{7cm}|}
    \hline
    You are a topic classifier. Given an input text you will classify it with zero or more of the following topics output in a JSON array format: \\
    Label: Citizen led initiatives to promote peace, Description: Citizen led initiatives to promote peace \\
    Label: Everything Fine, Description: People reporting that things are going smoothly. \\
    Label: Organization led Initiatives, Description: The police sometimes have community outreach events to promote peace and security during the polls. \\
    \hline
    \end{tabular}
    \caption{{\bf Prompt for \emph{Positive Events} tag prediction.}}
    \label{tab:prompt-positive-events}
\end{table}

\begin{table}[!th]
    \centering
    \small
    \tt
    \begin{tabular}{|p{7cm}|}
    \hline
    You are a topic classifier. Given an input text you will classify it with zero or more of the following topics output in a JSON array format: \\
    Label: Demonstrations, Description: Rallies and Marches \\
    Label: Mobilisation towards violence, Description: Situations where violence is imminent but has not started \\
    Label: Violent Attacks, Description: Attacks involving unarmed combatants, knives, or few (1 to 2) handguns \\
    Label: Abductions/kidnapping, Description: Acts of abductions, kidnapping, or hostage taking \\
    Label: Rumors, Description: Presence of a credible threat, along with unverified reports of violence, corruption, looting, etc. \\
    Label: Dangerous Speech, Description: Threats of violence \\
    Label: Vandalism and Physical Attacks on Property, Description: Vandalism and Physical Attacks on Property \\
    Label: Heavy police presence, Description: Presence of a large number of police officers \\
    \hline
    \end{tabular}
    \caption{{\bf Prompt for \emph{Security Issues} tag prediction.}}
    \label{tab:prompt-security-issues}
\end{table}

\begin{table}[!th]
    \centering
    \small
    \tt
    \begin{tabular}{|p{7cm}|}
    \hline
    You are a topic classifier. Given an input text you will classify it with zero or more of the following topics output in a JSON array format: \\
Label: voting irregularities, Description: Cases such as eligible voters being turned away or not allowed to vote, and ineligible voters allowed to vote \\
Label: voter integrity irregularities, Description: Cases such as importation of voters, voter impersonation, voter intimidation, bribing of voters, voters voting more than once, voter identification kit not working, etc. \\
Label: voter registration irregularities, Description: Issues with registration such as the register of voters missing, or voter names missing from the registry \\
Label: civic education gap, Description: Voter requesting or lacking crucial information about how to vote \\
Label: voter assistance irregularities, Description: Cases such as issues with proper identification, illiterate voters not being assisted, unusually many assisted voters, and the voter assister not taking the oath of secrecy \\
Label: alleged rigging, Description: Allegations of rigging elections \\
    \hline
    \end{tabular}
    \caption{{\bf Prompt for \emph{Voting Issues} tag prediction.}}
    \label{tab:prompt-voting-issues}
\end{table}


\clearpage
\section{Geotagging}
\label{sec:appendix-geotagging-prompts}

Figure \ref{fig:geotagging-has-location-prompt} shows the zero-shot prompt used to select the 2,818 reports for the geocoding evaluation.
This prompt is used to identify reports that contain locations in either their title or text.
Figure \ref{fig:location-extraction-prompt} shows the prompt used for the location extraction task along with the examples provided for in-context learning.

\begin{figure}[ht]
    \small
    \centering
    \begin{tabular}{|p{2.7in}|}
    \hline
     \texttt{You are an advanced text analysis model. Your task is to determine whether a given title and text contain a named location (e.g., a country such as Kenya, city such as Mombasa, landmark such as the Masaai Mara, or geographical region such as Nyanza). Named locations include any proper nouns that refer to places.}\\
     \texttt{Instructions:}\\
     \texttt{Input: You will be provided with a "Title" and "Text."}\\
    \texttt{Output: You should output "Yes" if you detect the presence of a named location in either the title or text, and "No" if there are no named locations.}\\
    \hline
    \end{tabular}
    \caption{\textbf{Prompt used to predict the presence of a location mention in a report.}}
    \label{fig:geotagging-has-location-prompt}
\end{figure}

\begin{figure*}[ht]
    \centering
    \small
    \begin{tabular}{|p{6in}|}
    \hline
     \texttt{You are a helpful, respectful and honest assistant. You are have been given the task of named location extraction. If you don't know the answer to a question, please don't share false information.}
    \texttt{Extract the exact match for all named location in the following text. Output a JSON object with the a single location field containing a list (e.g. \{``location": [``Nairobi", ``Nairobi County"]\})}.\\
    \\
     \texttt{Example 1:}\\
     \texttt{Title: ``IEBC Chairperson Wafula Chebukati Calls for Prayers for Staff Families"}\\
        \texttt{Text: ``Wafula Chebukati, the Independent Electoral and Boundaries Commission (IEBC) chairperson in Bomas of Kenya, has appealed to Kenyans to pray for the spouses and children of his staff."}\\
        \texttt{Output: \{``label": [``Bomas of Kenya"]\}}
    \\\\
        \texttt{Example 2: 
        Title: ``Nairobi Voters Speak Out: A Look at the 2022 Kenyan General Elections"
        Text: "As the 2022 Kenyan general elections approach, voters in the capital are eagerly awaiting their chance to make their voices heard. With a diverse range of candidates and issues on the ballot, the city's residents are poised to play a crucial role in shaping the future of the country."
        Output: {``label": ["Nairobi"]}}
    \\\\
         \texttt{Example 3:}\\ 
        \texttt{Title: "The Future of Agriculture: A Look at the 2022 Kenyan General Elections"}\\
        \texttt{Text: "As the 2022 Kenyan general elections approach, the future of agriculture is a top concern for many voters. With issues such as land use, irrigation, and crop yields at the forefront of the campaign, we will be hosting events in Nakuru, Eldoret and Nyeri to take a closer look at some of the candidates' positions on these critical issues."}\\
        \texttt{Output: {"label": ["Nakuru", "Eldoret", "Nyeri", "Nakuru, Eldoret and Nyeri"]}}\\\\

        \texttt{Example 4:}\\
        \texttt{Title: "Bomet Teachers Training College Students Endorse Local Candidate in 2022 Kenyan General Elections"}\\
        \texttt{Text: "At Bomet Teachers Training College, Kaplong-Narok-Maai Road located in Bomet in the Rift Valley, students have come together to endorse a local candidate in the upcoming general elections. The candidate, who has been actively involved in community service, has gained the support of the students due to his commitment to improving the lives of the residents."}\\
        \texttt{Output: \{``label": ["Bomet Teachers Training College", "Bomet Teachers Training College, Kaplong-Narok-Maai Road", "Kaplong-Narok-Maai Road", "Bomet", "Rift Valley", "Bomet Teachers Training College, Kaplong-Narok-Maai Road located in Bomet in the Rift Valley"]\}}\\\\
        
        \texttt{Example 5:}\\
        \texttt{Title: "Manyatta Estate Residents Demand Better Representation in Upcoming Elections"}\\
        \texttt{Text: "Residents of Manyatta Estate in Kisumu, Western Kenya, are calling for better representation in the upcoming 2022 Kenyan general elections. With a growing population and a lack of basic amenities, the community is eager to have their voices heard and their needs addressed."}\\
        \texttt{Output: {``label": ["Manyatta Estate", "Kisumu, Western Kenya", "Kisumu", "Western Kenya"]}}\\ \\

        \texttt{Example 6:}\\
        \texttt{Title: "Mwingi MP Victor Munyasia Hosts Public Debate Ahead of 2022 Elections"}\\
        \texttt{Text: "Mwingi MP Victor Munyasia recently hosted a public debate for residents of Mwingi and surrounding areas to discuss issues affecting the community and their hopes for the 2022 Kenyan general elections."}\\
        \texttt{Output: {``label": [``Mwingi"]}}\\
        \\
        \texttt{Example 7:}\\
        \texttt{Title: "Busy Polling Station"}\\
        \texttt{Text: "Hey! Just got to the Westlands Primary School polling station in Westlands Constituency. The place is packed, lines are super long but everyone seems determined. People are chatting, sharing snacks, and even singing. It's a great vibe. Voting is slow but steady. Make sure you bring water and a hat if you're coming later. \#KenyaDecides2022"}\\
        \texttt{Output: \{``label": [``Westlands Primary School'', "Westlands Primary School polling station", "Westlands", "Westlands Constituency", "Westlands Primary School polling station in Westlands Constituency"]\}}\\
        \\
        
    \texttt{New task:}\\
 \texttt{Title: ``Urging The Youth To Promote Peace And Liberty"} \\
\texttt{Text: At PrideInn Paradise Mombasa encouraging YOUTH to promote peace and liberate themselves as well as the society at large with an aim to prevent a repeat of the 2007 PEV \#UchaguziWaAmani @YEDNetworkKe @VybezRadioKE \#Vijanapeace} \\
\texttt{Output:}\\
    \hline
    \end{tabular}
    \caption{\textbf{Prompt for location extraction.}}
    \label{fig:location-extraction-prompt}
\end{figure*}

\section{License}
\label{sec:appendix-license}
The dataset is available for research purposes only, upon submission and review of a data request form.


\end{document}